\pdfoutput=1

\documentclass[11pt]{article}

\usepackage[final]{acl}

\usepackage{times}
\usepackage{latexsym}
\usepackage[T1]{fontenc}
\usepackage[utf8]{inputenc}
\usepackage{microtype}

\usepackage{algorithm}
\usepackage{algpseudocode}
\usepackage{graphicx}
\usepackage{svg}
\usepackage{amsmath}
\usepackage{float}
\usepackage{caption}
\usepackage{subcaption}
\usepackage{soul}
\usepackage{multirow}
\usepackage{multicol}
\usepackage{numprint}
\usepackage{longtable}
\usepackage{makecell}
\usepackage{tikz}
\usepackage{booktabs, multirow} % for borders and merged ranges
\usepackage{changepage, threeparttable} % for wide tables
\usepackage{xcolor}
\usepackage{enumitem}
\usepackage{lipsum}

\definecolor{GoodGreen}{RGB}{3, 146, 94}
\definecolor{BadRed}{RGB}{250, 70, 70}

\graphicspath{ {./images/} }

\title{Optimizing Negative Prompts for Enhanced Aesthetics and Fidelity in Text-To-Image Generation}

\author{Michael Ogezi and Ning Shi \\
  Department of Computing Science, \\ 
  University of Alberta, Edmonton, Canada \\
  \href{mailto:ogezi@ualberta.ca}{\color{black}\texttt{\{mikeogezi,ning.shi\}@ualberta.ca}} \\
}

\begin{document}
\maketitle

\begin{abstract}
In text-to-image generation, using negative prompts, which describe undesirable image characteristics, can significantly boost image quality. 
However, producing good negative prompts is manual and tedious. 
To address this, we propose NegOpt, a novel method for optimizing negative prompt generation toward enhanced image generation, using supervised fine-tuning and reinforcement learning. 
Our combined approach results in a substantial increase of 25\% in Inception Score compared to other approaches and surpasses ground-truth negative prompts from the test set. 
Furthermore, with NegOpt we can preferentially optimize the metrics most important to us. 
Finally, we detail the construction of \href{https://huggingface.co/datasets/mikeogezi/negopt_full}{Negative Prompts DB}, a publicly available dataset of negative prompts.
\end{abstract}

\section{Introduction}

% - what is the problem
Prompting has become the standard approach for adapting large models to specific tasks \cite{liu2023pre}, with prompt optimization techniques \cite{lester2021power, deng2022rlprompt, li2021prefix, liu-etal-2022-p} playing a crucial role in enhancing performance. This trend is also evident in text-to-image generation \cite{ramesh2022hierarchical}. Despite recent improvements in image-generation quality \cite{rombach2022high}, challenges remain. In this work, we look to enhance image-generation aesthetics and fidelity by optimizing negative prompts.

\begin{figure}[t!]
    \centering
    \begin{tabular}{@{\hspace{1mm}}c@{\hspace{1mm}}c@{\hspace{1mm}}|@{\hspace{1mm}}c@{\hspace{1mm}}c@{\hspace{1mm}}}
        \small {\textsc{NegOpt}} & \small {{Baseline}} & \small {\textsc{NegOpt}} & \small {{Baseline}} \\
        \includegraphics[width=0.225\linewidth]{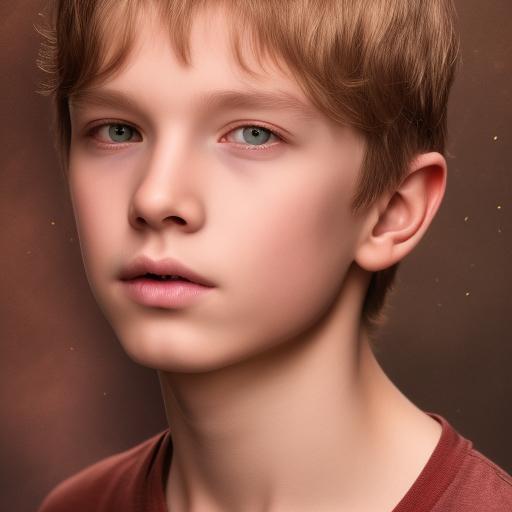} & \includegraphics[width=0.225\linewidth]{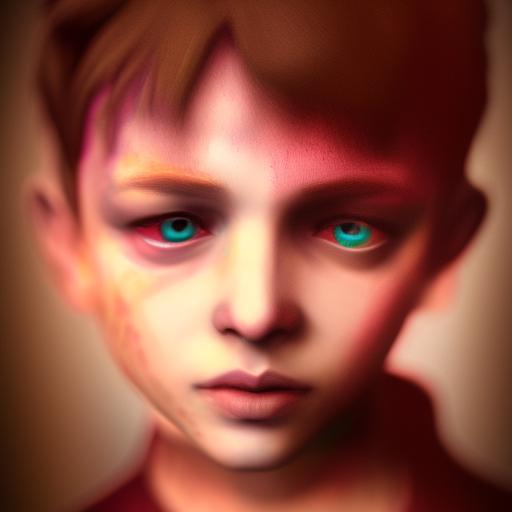} & \includegraphics[width=0.225\linewidth]{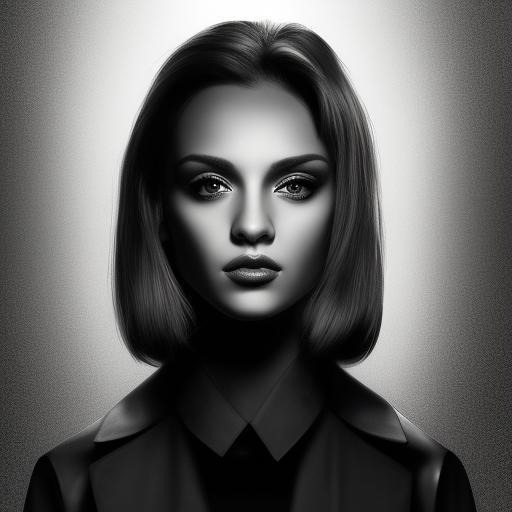} & \includegraphics[width=0.225\linewidth]{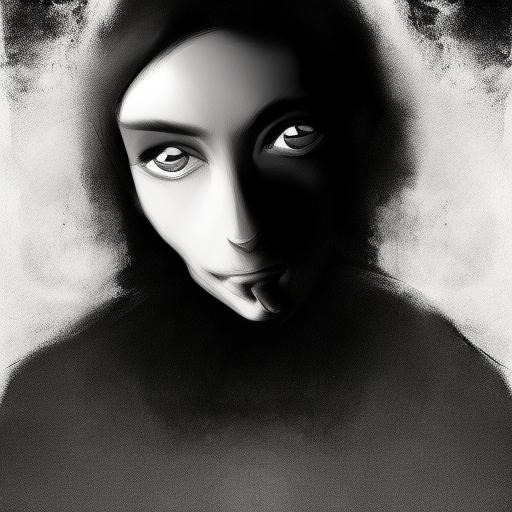} \\
    \end{tabular}
    \caption{Images generated with \textsc{NegOpt} 
    (\textsc{SFT+RL} specifically) 
    negative prompts vs. {baseline} images.}
    \label{fig:ours_vs_baseline}
\end{figure}

% - why is the problem important
Images produced by text-to-image generation models \cite{ramesh2022hierarchical, rombach2022high} sometimes suffer from issues such as blurriness and poor framing. 
Negative prompts can divert models from such unwanted characteristics. \citet{Wong_2023} shows that negative prompts with phrases like \textit{``blurriness''} and  \textit{``out of frame''}, can be more effective than \textit{normal} prompts with phrases such as \textit{``clear''} and \textit{``good framing''} in preventing unwanted characteristics (see Figure \ref{fig:ours_vs_baseline}).

% - how other people deal with the problem
Recent work has attempted to improve image quality by optimizing \textit{normal} prompts \cite{hao2022optimizing}. Prior to this, however, optimizing prompts for image generation was performed manually in a tedious and error-prone process.

% - what is our great idea of dealing with the problem
To address these issues, we propose a novel approach called NegOpt, which performs \textbf{neg}ative prompt \textbf{opt}imization through a two-step process. First, we treat the task of generating prompts as a sequence-to-sequence (seq2seq) language modeling problem, where the input is a \textit{normal} prompt, $p$ and the output is a negative prompt, $p'$. We fine-tune a seq2seq model on a first-of-its-kind dataset of $p$-$p'$ pairs, which we construct and call Negative Prompts DB. Second, we frame the task as a reinforcement learning problem, maximizing a scalar reward signal based on the quality of images generated with $p$ and $p'$.

\begin{figure*}[t!]
    \centering
    \includegraphics[width=\linewidth]{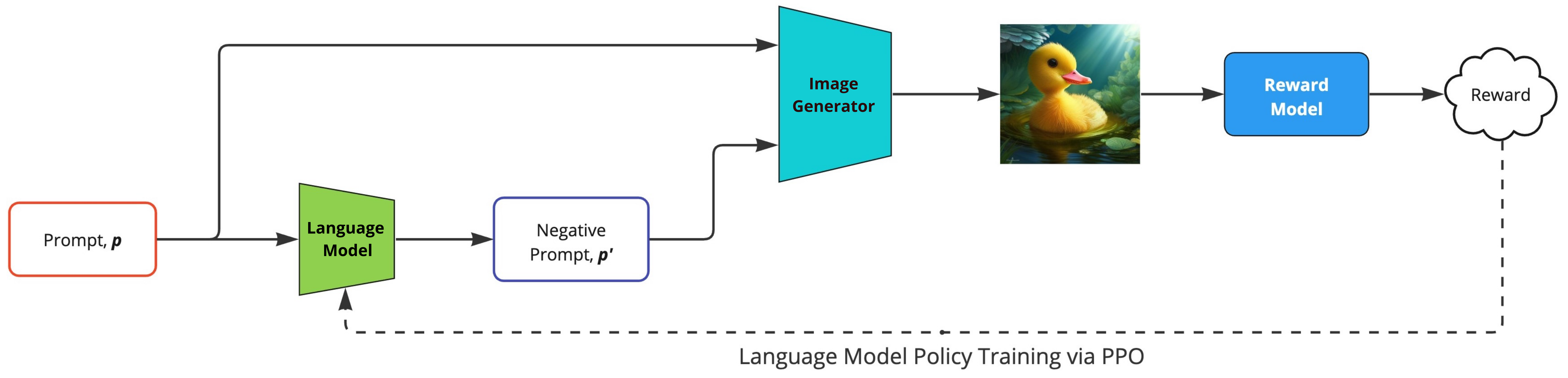}
    \caption{
    In NegOpt, we first use a fine-tuned sequence-to-sequence language model to generate a negative prompt, $ p' $, given a \textit{normal} prompt, $ p $.
    Next, we use $ p $ and $ p' $ to generate an image with an image generator. 
    Finally, we further optimize our language model based on the reward received for the generated image.
    }
    \label{fig:method}
\end{figure*}

% - what are our results briefly
Our results demonstrate a significant improvement in generated-image aesthetics and fidelity with a {$ \sim $}25\% increase in Inception Scores as well as improvements in other metrics. We even surpass the ground-truth negative prompts when evaluating on the test set. Moreover, we show that NegOpt improves our most valued metrics, which are aesthetics and fidelity, without sacrificing others such as prompt alignment. Our key contributions are:

\begin{enumerate}
    \item We introduce NegOpt, a method for optimizing negative prompts for image generation that improves aesthetics and fidelity.
    
    \item We construct Negative Prompts DB, a dataset specifically aggregating negative prompts.
\end{enumerate}

% - what is the structure of the paper
% The structure of the paper is as follows: Section \ref{sec:background} provides background on image generation, prompting, and prompt optimization; Section \ref{subsec:dataset} details our constructed dataset, Negative Prompts DB; Section \ref{sec:negopt} presents our NegOpt method; Section \ref{sec:experiments} presents the experimental setup and results; Section \ref{sec:results} discusses the implications of our findings; and Section \ref{sec:conclusion} concludes the paper and suggests future research directions.

\section{Background}
\label{sec:background}

% \paragraph{Image Generation via Diffusion Models}
% Diffusion Models (DMs) are a class of generative models that excel at image generation tasks \cite{dhariwal2021diffusion}. They use a progressive denoising technique for corrupted images, offering a simpler training approach compared to predecessors like Generative Adversarial Networks \cite{goodfellow2020generative}. Latent DMs denoise a lower-dimensional noisy image and then convert it to the final image \cite{rombach2022high}. In our study, we employ Stable Diffusion, a conditional latent DM variant that typically uses text prompts for conditioning.

Recent research has made significant progress in optimizing prompts for various tasks, mainly focusing on gradient-based methods \cite{qin2021learning, li2021prefix, liu-etal-2022-p} and reinforcement learning approaches \cite{deng2022rlprompt}. However, most of this work targets pure natural language processing tasks.

In text-to-image generation, one study has successfully explored optimizing \textit{normal} prompts \cite{hao2022optimizing}, but negative prompt optimization remains unaddressed. This gap is particularly noteworthy considering the extensive use of negative prompts in online communities\footnote{\url{https://stable-diffusion-art.com/how-to-use-negative-prompts/}}.

\begin{table*}[t!]
  \small
  \centering
  \begin{tabular}{>{\centering\arraybackslash}m{0.22\textwidth}
                  >{\centering\arraybackslash}m{0.04\textwidth}
                  >{\centering\arraybackslash}m{0.32\textwidth}
                  >{\centering\arraybackslash}m{0.09\textwidth}
                  >{\centering\arraybackslash}m{0.21\textwidth}}
    \toprule
    {Prompt} & & {Negative Prompt} & {Image} & {Other Images} \\
    \midrule
    \multirow{3}{4cm}[-7.25ex]{A Wolf, high detailed fur, monochrome charcoal painting art, detail work, various tones of black, blended, colorful detailed eyes, by Gerd Amble, 8k, 2d render}
    & {\rotatebox[origin=l]{90}{\textsc{None}}}
    & -
    & \includegraphics[width=0.1\textwidth]{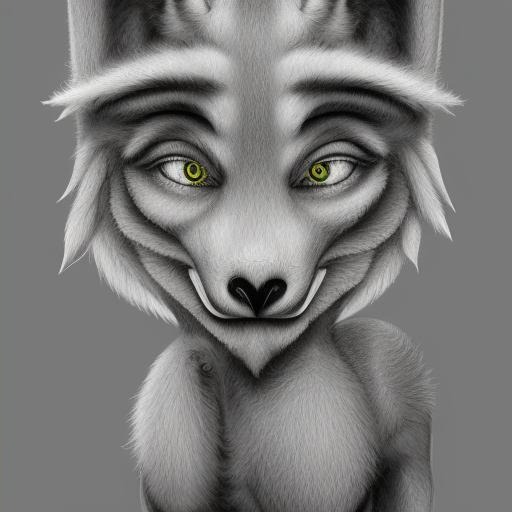} & \includegraphics[width=0.1\textwidth]{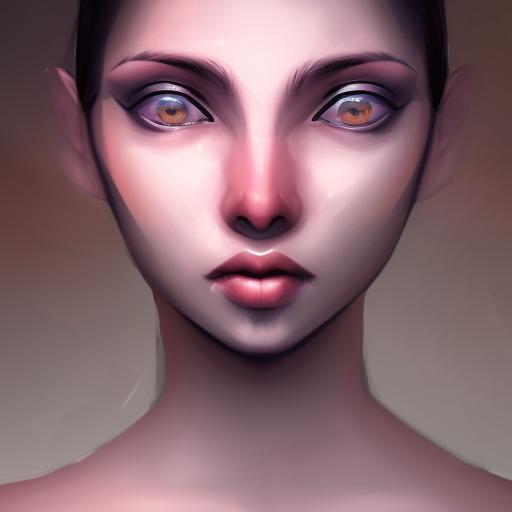} \includegraphics[width=0.1\textwidth]{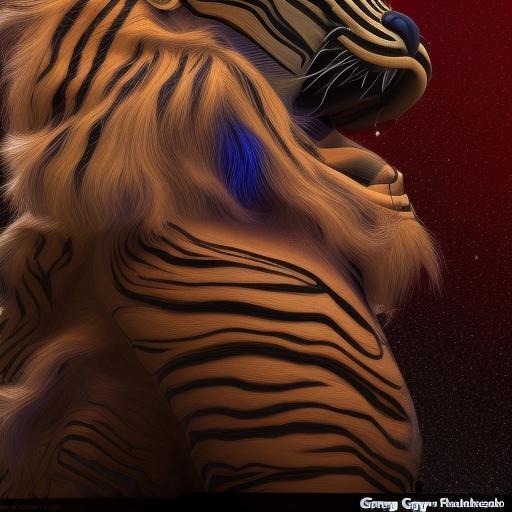} \\
    & {\rotatebox[origin=l]{90}{\textsc{\thead{Ground\\Truth}}}}
    & multiple heads, multiple wolfs, close up
    & \includegraphics[width=0.1\textwidth]{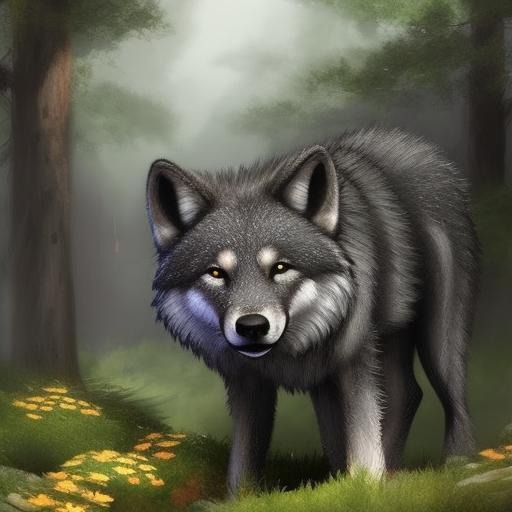} & \includegraphics[width=0.1\textwidth]{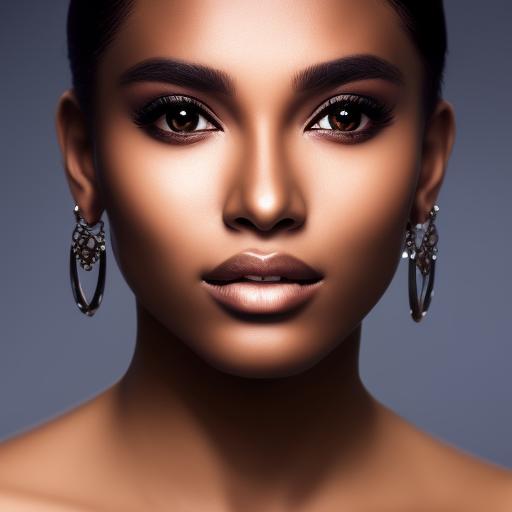} \includegraphics[width=0.1\textwidth]{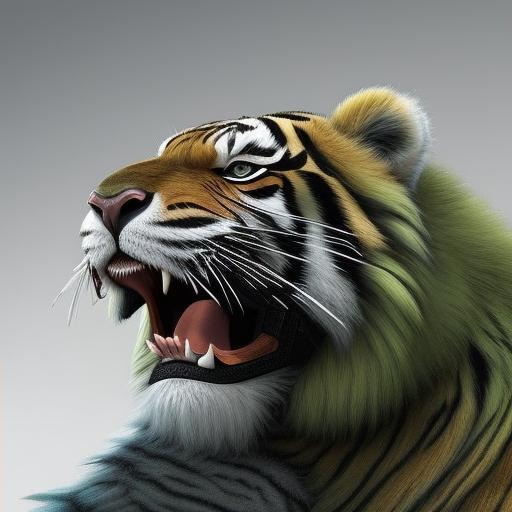} \\
    & {\rotatebox[origin=l]{90}{\textsc{\thead{NegOpt}}}}
    & elongated body, cropped image, red eyes, out of frame, draft, deformed hands, signatures, big hair, twisted fingers, double image, long neck, malformed hands, multiple heads, extra limb, ugly...
    % elongated body, cropped image, red eyes, out of frame, draft, deformed hands, signatures, big hair, twisted fingers, double image, long neck, malformed hands, multiple heads, extra limb, ugly, poorly drawn hands, missing limb, disfigured, cut-off, kitsch, ugly, over saturated, grain, low-res, Deformed, blurry, bad anatomy, disfigured, poorly drawn face, mutation, mutated, floating limbs, disconnected limbs, out of focus, long body, disgusting, poorly drawn,
    & \includegraphics[width=0.1\textwidth]{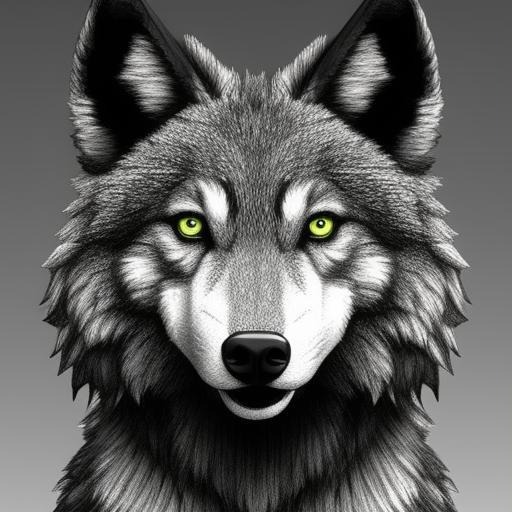} & \includegraphics[width=0.1\textwidth]{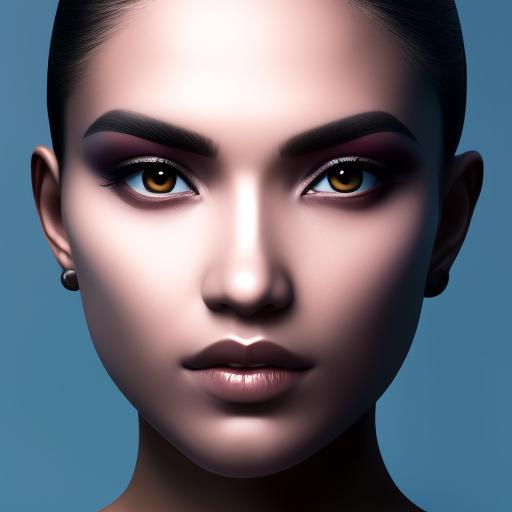} \includegraphics[width=0.1\textwidth]{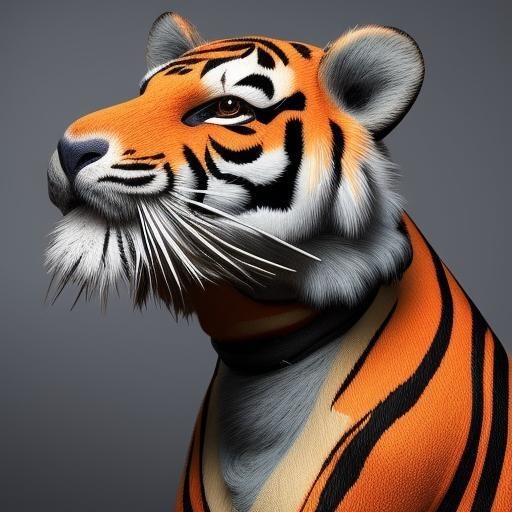} \\
    \bottomrule
  \end{tabular}
  \caption{
  An example of the qualitative improvement in fidelity and aesthetics with \textsc{NegOpt} (\textsc{SFT+RL} specifically) and \textsc{Ground Truth}, both using negative prompts vs. \textsc{None} which is without negative prompts.
  The shown prompt only corresponds to the first image column. See Table \ref{table:appendix_examples_1} in the appendices for more examples.
  }
  \label{table:descriptions_comparison}
\end{table*}

\section{NegOpt}
\label{sec:negopt}

Our contributions consist of a unique dataset and a novel method for optimizing negative prompts.

\subsection{Dataset: Negative Prompts DB}
\label{subsec:dataset}

We compiled a collection of publicly-available prompt samples from Playground AI\footnote{\url{https://playgroundai.com/}}, selecting posts submitted between December 1, 2022, and March 7, 2023, that received at least one like. This resulted in a total of 256,224 samples, collected on March 9, 2023. The following attributes are available: {\small \texttt{id}}, {\small \texttt{created\_at}}, {\small \texttt{model}}, {\small \texttt{sampler}}, {\small \texttt{likes}}, {\small \texttt{height}}, {\small \texttt{steps}}, {\small \texttt{width}}, {\small \texttt{cursor}}, {\small \texttt{url}}, and {\small \texttt{cfg\_scale}}. 
In our experimental setup, we describe how we use subsets of Negative Prompts DB within our experiments (see Section \ref{par:sft_subset} and Section \ref{par:rl_subset}).

\subsection{Core Method: NegOpt}
\label{subsec:method}

NegOpt consists of two phases: Supervised Fine-Tuning (SFT) and Reinforcement Learning (RL) toward producing optimized negative prompts.

\subsubsection{Phase 1: Supervised Fine-tuning (SFT)}

In this phase, we fine-tune a sequence-to-sequence (seq2seq) model \cite{sutskever2014sequence} using a subset of the Negative Prompts DB dataset (see Section \ref{par:sft_subset}). 
% We select seq2seq because our negative prompts are a transformation of our normal prompts.
We set the \textit{normal} prompt as the \textit{source}, and the \textit{negative} prompt as the \textit{target}.
The fine-tuned model then serves as a foundation for the subsequent RL phase.

\subsubsection{Phase 2: Reinforcement Learning (RL)}

In this phase, we preferentially optimize the fine-tuned model's performance with RL. As depicted in Figure \ref{fig:method}, we train the model to maximize a reward, corresponding to the quality of the generated image, $i$. We define the reward, $r$ as a weighted sum of three terms:
\begin{equation}
\begin{gathered}
    r(p, i) := \alpha \cdot s_{aesthetics}(i) \\ + \beta \cdot s_{alignment}(p, i) + \gamma \cdot s_{fidelity}(i)
\end{gathered}
\label{eqn:reward}
\end{equation}

Where:

\begin{itemize}[itemsep=-1pt]
    \item $\alpha$, $\beta$, and $\gamma$ are coefficients denoting the relative importance of $s_{aesthetics}$, $s_{alignment}$, and $s_{fidelity}$ (see following points) respectively.
    
    \item $s_{aesthetics}$ is an aesthetics score obtained from a fully connected neural network trained to take CLIP \cite{radford2021learning} image embeddings as input and predict a score representing human aesthetic preferences\footnote{\url{https://github.com/christophschuhmann/improved-aesthetic-predictor}}.

    \item $s_{alignment}$ is an alignment score determined by the cosine similarity between CLIP embeddings of the \textit{normal} prompt and the generated image.

    \item $s_{fidelity}$ is a fidelity score based on Inception Scores \cite{salimans2016improved}.
\end{itemize}

\section{Experimental Setup}
\label{sec:experiments}

We use T5\footnote{\url{https://hf.co/t5-small}} \cite{raffel2020exploring} as our seq2seq language model and Stable Diffusion\footnote{\url{https://hf.co/stabilityai/stable-diffusion-2-base}} \cite{rombach2022high} as our image generator.

\subsection{SFT}

\paragraph{SFT Dataset Subset}
\label{par:sft_subset}

We choose a subset of the Negative Prompts DB dataset, selecting only Stable Diffusion \cite{rombach2022high} posts with 20+ likes. This subset consists of 5,790 samples. The train/validation/test split ratio is 90:5:5.

\paragraph{SFT Training}
After hyperparameter search, we select the Adam optimizer \cite{kingma2014adam} with a learning rate of $1\mathrm{e}{-3}$, $0.01$ weight decay, $256$ warm-up steps, and an effective batch size of $32$. We fine-tune for $16$ epochs with the prefix instruction \textit{``generate a negative prompt for:''}.

\subsection{RL}

\paragraph{RL Dataset Subset}
\label{par:rl_subset}

We use a more selective subset of 466 samples with 100+ likes from the SFT train and validation splits. We set the new train/validation split ratio to 90:10.

\paragraph{RL Training}
\label{par:rl_training}

After hyperparameter search, we choose the Adam optimizer with a learning rate of $1\mathrm{e}{-5}$ and an effective batch size of $16$. We train for $8$ epochs and use greedy decoding during evaluation by setting the temperature to $0$. We prioritize aesthetics by setting $ \alpha = 5 $, $ \beta = 1 $, and $ \gamma = 1 $. We employ Proximal Policy Optimization (PPO) \cite{schulman2017proximal}, a policy gradient algorithm that updates the model's policy using gradient descent.

\subsection{Evaluation}

\paragraph{Image Generation Details}
We run Stable Diffusion for $25$ steps with a guidance scale of $7.5$ on the prompts from the SFT test split. We generate images with 8 different seeds and record the mean of the evaluation metrics.

\paragraph{Evaluation Metrics} We employ these metrics:

\begin{itemize}[itemsep=-1pt]
    \item Inception Score \cite{salimans2016improved}: This assesses both the quality and diversity of the generated images.
    
    \item CLIP Score \cite{hessel2021clipscore}: This measures the semantic similarity between images and text, with scores ranging from 0 to 100.
    
    \item Aesthetics Score\footnote{\href{https://github.com/christophschuhmann/improved-aesthetic-predictor}{\url{github.com/christophschuhmann/improved-aesthetic-predictor}}}: This gauges the aesthetic appeal of images based on human preferences, with scores ranging from 0 to 10.
    
    \item Human Evaluation: We have evaluators rank generated images (see Appendix \ref{app:human_evaluation}) then compute the mean rank for each method. Scores range from 1 to the number of methods being evaluated (6, in this case).
\end{itemize}

\paragraph{Baselines, Previous Methods, \& Method Variants}
\label{par:previous_methods}

We compare to two baselines: \textsc{None} and \textsc{Ground Truth}, one previous method: \textsc{Promptist}, and three \textsc{NegOpt} variants: \textsc{SFT}-only, \textsc{RL}-only, and \textsc{SFT}+\textsc{RL}.
We describe them next:

\begin{itemize}[itemsep=-1pt]
    \item \textsc{None} does not utilize a negative prompt.
    % \item \textsc{Rand}: This approach uses a randomly sampled negative prompt.
    % \item \textsc{Fixed}: This approach applies the same optimized negative prompt for all instances.
    \item \textsc{Ground Truth} uses the ground-truth negative prompt from the dataset.
    \item \textsc{Promptist} \cite{hao2022optimizing} trains a policy on to augment \textit{normal} prompts. 
    \item \textsc{SFT}-only uses the negative prompt from a model that is trained with SFT but not RL.
    \item \textsc{RL}-only uses the negative prompt from a model trained with RL but not SFT.
    \item \textsc{SFT+RL} combines both SFT and RL.
\end{itemize}

\begin{table}[t!]
\def\arraystretch{1}
\setlength{\tabcolsep}{5pt}
\small
\centering
\begin{tabular}{@{}cccccc@{}}
\toprule
\multicolumn{2}{c}{\multirow{3}{*}{}} & \multicolumn{4}{c}{Metrics} \\

\cmidrule(lr){3-6} \multicolumn{2}{c}{} & \thead{Inception\\Score\\(↑)} & \thead{CLIP\\Score\\(↑)} & \thead{Aesthetics\\Score\\(↑)} & \thead{Mean\\Human\\Rank\\(↓)} \\

\midrule
\multirow{4}{*}{\rotatebox[origin=c]{90}{Baselines}} & {\textsc{None}} & 5.58 & \textbf{32.16} & 6.08 & 3.42 \\

\cmidrule{2-6}
& {\textsc{\thead{Ground\\Truth}}} & 6.82 & \underline{30.98} & \underline{6.28} & 2.96 \\

\midrule
\multicolumn{2}{r}{\;\;\;\textsc{\;\;Promptist}} & 6.61 & 20.49 & 5.23 & 4.96 \\

\midrule
\multirow{4}{*}{\rotatebox[origin=c]{90}{\textsc{NegOpt}}} & \multicolumn{1}{l}{\;\;\:\textsc{RL}-only} & 6.98 & 28.37 & 5.76 & 3.94 \\

\cmidrule{2-6}
& {\textsc{SFT}-only} & \textbf{7.16} & 30.58 & \underline{6.28} & \textbf{2.82} \\

\cmidrule{2-6}
& {\textsc{SFT+RL}} & \underline{7.08} & {30.88} & \textbf{6.30} & \underline{2.90} \\

% \midrule
% \multicolumn{2}{c}{\Delta (\%)} & \color{GoodGreen}{+26.88 $} & \color{BadRed}{-3.98 $} & \color{GoodGreen}{+3.49 $} \\
\bottomrule
\end{tabular}
\caption{
The relative performance of baselines, previous methods, and method variants (see Section \ref{par:previous_methods} for details) on the test set.
% $ \Delta $ refers to the delta between \textsc{NegOpt} and \textsc{None}.
The \textbf{best} is emboldened, the \underline{second best} is underlined, and ↑/↓ indicates that higher/lower is better.
}
\end{table}

\section{Results and Discussion}
\label{sec:results}

\subsection{Results}

Our results, presented in Table \ref{sec:results}, demonstrate the effectiveness of NegOpt compared to the baselines and alternative methods discussed in Section \ref{par:previous_methods}. In terms of Inception Score, our fidelity measure, we achieve substantial improvements of 24.8\% and 8.0\% over the \textsc{None} baseline and the top previous method, \textsc{Promptist}, respectively. For the Aesthetics Score, our aesthetics measure, we outperform \textsc{None} and \textsc{Promptist} by 3.5\% and 18.6\% respectively. Although our CLIP Score, the alignment measure, drops by 4.1\% compared to \textsc{None}, this is acceptable as it is our least important concern and the decrease is relatively small. We also observe that \textsc{RL}-only performs worse than \textsc{SFT}-only and to a greater extent \textsc{SFT+RL}.

\subsection{Discussion}

\paragraph{SFT provides a strong start}
The \textsc{SFT}-only approach already demonstrates a significant improvement in image quality, even surpassing NegOpt in Inception Score. This validates the quality of our dataset and the initial part of our approach. However, there is more to consider.

\paragraph{Learning better negative prompts than ground-truth}
Interestingly, NegOpt learns better negative prompts than the ground truth in the test set. This is not surprising, as the negative prompts were created by humans, who are essentially performing some manual optimization task. Consequently, NegOpt can leverage patterns in the training data more efficiently than those who produced it. Next, we discuss the importance of RL.

\paragraph{RL optimizes for metrics we care more about}
By incorporating RL, we can better control our performance on specific metrics. Based on our settings of $ \alpha $, $ \beta $, and $ \gamma $ from Section \ref{par:rl_training}, we prioritized the aesthetics. As a result, \textsc{SFT+RL} outperforms \textsc{SFT}-only in aesthetics. Thus, the addition of RL, and more specifically a multi-component reward function (see Equation \ref{eqn:reward}), enables us to tailor image generation toward what we value most without any substantial degradation in the less valued ones.

\section{Conclusion}
\label{sec:conclusion}

In this paper, we introduced NegOpt, a novel method for optimizing negative prompts for text-to-image generation tasks. To the best of our knowledge, this is the first approach specifically targeting the optimization of negative prompts. Additionally, we constructed Negative Prompts DB, the first dataset designed for negative prompts. NegOpt demonstrates strong performance in terms of the evaluation metrics: Inception Score, CLIP Score, and Aesthetics Score, showcasing its effectiveness in enhancing image generation. We showed that Supervised Fine-Tuning provides a solid foundation, while further optimization through Reinforcement Learning allows for more targeted improvements in the most relevant metrics. On a broader level, our work highlights the potential of expressing optimization problems in source modalities, such as language, to achieve significant performance gains in target modalities, such as vision. This opens up new avenues for future research in optimizing prompts for non-language tasks.

\section*{Acknowledgements}

We thank Logan Woudstra for helping out with human evaluation and Lili Mou for providing supplementary computing resources.
Finally, we thank Greg Kondrak for general guidance in our research.

\section*{Ethics Statement}
\label{sec:ethics}

In developing our NegOpt method for optimizing negative prompts in image generation, we are committed to upholding ethical principles and ensuring responsible use. We acknowledge potential ethical concerns, including dataset bias, potential misuse, privacy concerns, ethical reinforcement learning, and transparency and explainability. To address these concerns, we strive to:

\begin{enumerate}[itemsep=-1pt]
    \item Assess and mitigate biases in the data, promoting fairness and inclusivity.
    \item Encourage responsible use by providing guidelines and recommendations.
    \item Respect user privacy by adhering to data protection regulations.
    \item Explore incorporating ethical considerations into the reinforcement learning reward function.
    \item Improve the transparency and explainability of our method.
\end{enumerate}

We believe that addressing these ethical concerns contributes to the development of ethically sound and beneficial image-generation technology. We encourage future research to consider these ethical aspects and collaborate in fostering responsible innovation in the field.

\section*{Limitations}
\label{sec:limitations}

\paragraph{Dataset bias} 
The Negative Prompts DB dataset may contain inherent biases, which could lead to biased image generation outcomes.

\paragraph{Potential misuse}
Our method could be misused to generate harmful content or misinformation, emphasizing the need for guidelines and safeguards.

\paragraph{Privacy concerns}
Although we do not use personally identifiable information, there may still be privacy concerns related to using user-generated content from Playground AI.

\paragraph{Ethical reinforcement learning}
The reward function in our method may not fully capture ethical implications, necessitating the exploration of incorporating ethical considerations.

\paragraph{Transparency and explainability}
The deep learning models used in our method may raise concerns about the interpretability of the generated negative prompts and resulting images.

\bibliography{anthology,custom}
\bibliographystyle{acl_natbib}

\appendix

\newpage
\section*{Appendix}
\label{sec:appendix}

\section{Human Evaluation}
\label{app:human_evaluation}

The instruction provided to human evaluators is: ``For each sample, you have six images and a prompt. You are to rank all images by two criteria of equal importance: (1) Beauty/aesthetic pleasantness, and (2) Relevance to the prompt. The ranking should be an ordered list of numbers (eg. 1,5,6,4,2,3; if you think the first image is the best and the third image is the worst).''

\section{NegOpt Examples}
\label{app:negopt_examples}

See Table \ref{table:appendix_examples_1} for NegOpt vs. all baselines and methods.

\begin{table*}[t]
  \small
  \centering
  \begin{tabular}{>{\centering\arraybackslash}m{0.25\textwidth}
                  >{\centering\arraybackslash}m{0.05\textwidth}
                  >{\centering\arraybackslash}m{0.35\textwidth}
                  >{\centering\arraybackslash}m{0.10\textwidth}}
    \toprule
    {Prompt} & & {Negative Prompt} & {Image} \\
    \midrule
    
    \multirow{6}{4cm}[-24.25ex]{lushill style, Woman head and shoulders portrait. style by Sir John Tenniel, Beatrix Potter, Brian Froud, Enki Bilal, Pascal Campion. Lovely flower ornaments on the wall behind her.} 
    & {\rotatebox[origin=l]{90}{\textsc{None}}}
    & -
    & \includegraphics[width=0.1\textwidth]{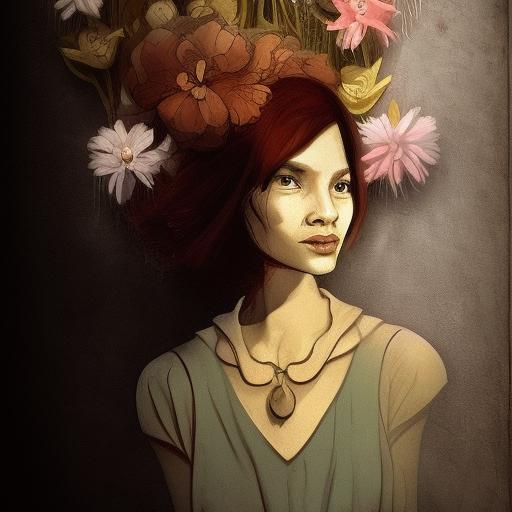} \\
    
    & {\rotatebox[origin=l]{90}{\textsc{\thead{Ground\\Truth}}}}
    & open-mouth, hands, duplicates, blur haze
    & \includegraphics[width=0.1\textwidth]{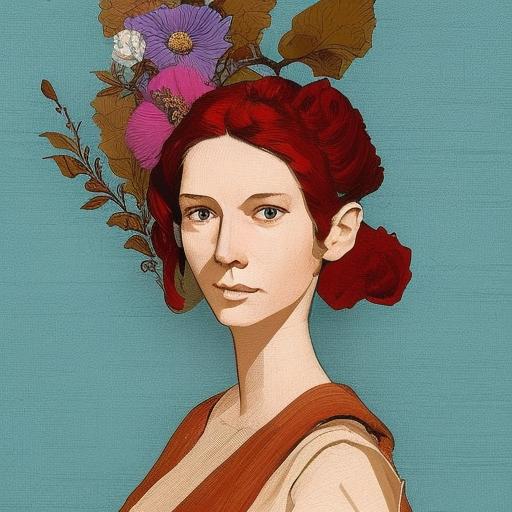} \\
    
    & {\rotatebox[origin=l]{90}{\textsc{Promtist}}}
    & -
    & \includegraphics[width=0.1\textwidth]{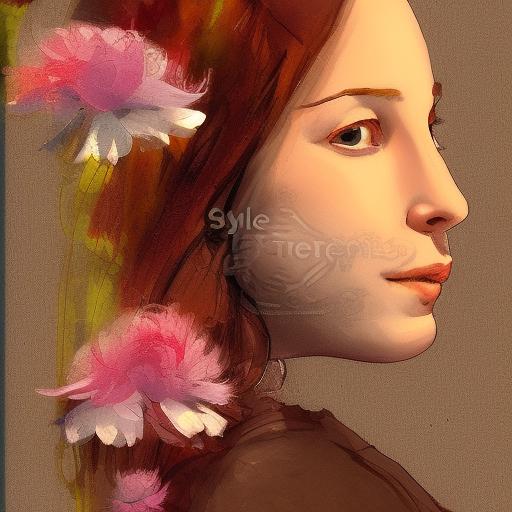} \\

    & {\rotatebox[origin=l]{90}{\textsc{RL}-only}}
    & , Brian Froud, Enki Bilal, Pascal Campion.
    & \includegraphics[width=0.1\textwidth]{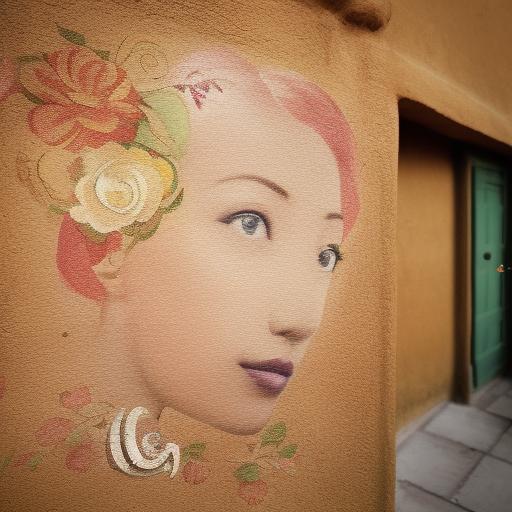} \\
    
    & {\rotatebox[origin=l]{90}{\textsc{SFT}-only}}
    & elongated body, cropped image, red eyes, out of frame, draft, deformed hands, signatures, big hair, twisted fingers, double image, long neck, malformed hands, multiple heads, extra limb, ugly...
    & \includegraphics[width=0.1\textwidth]{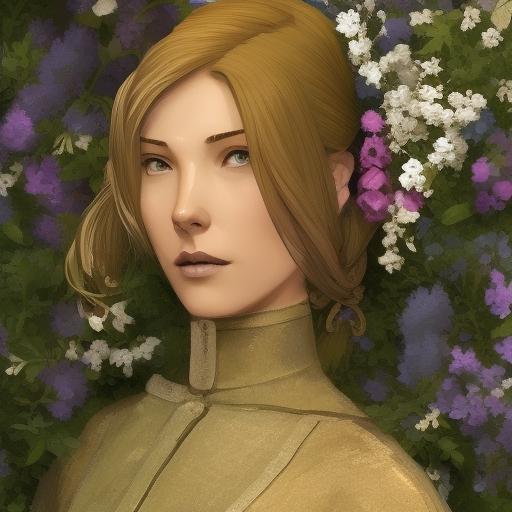} \\
    
    & {\rotatebox[origin=l]{90}{\textsc{SFT+RL}}}
    & elongated body, cropped image, red eyes, out of frame, draft, deformed hands, signatures, big hair, twisted fingers, double image, long neck, malformed hands, multiple heads, extra limb, ugly...
    & \includegraphics[width=0.1\textwidth]{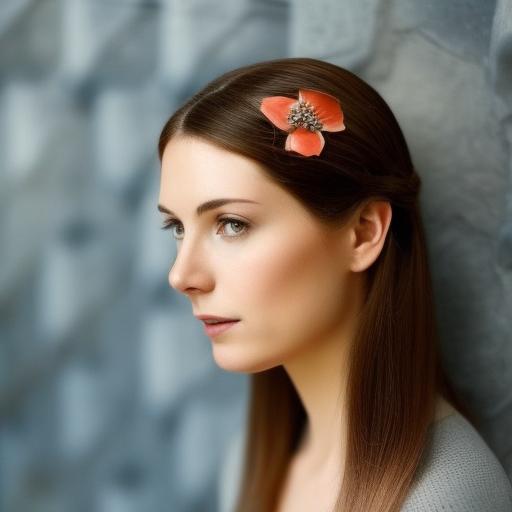} \\

    \midrule

    \multirow{6}{4cm}[-24.25ex]{elven architecture a small cottage intricate detailed wonder ancient woods background spring flowers vines,  beautiful garden with brook in background, golden sunlight and sunbeams photographic render, smooth sharp focus...}
    % hyper detailed, intricate detailed, hyper realistic, exceptional quality, brilliant realistic cinematic colour 4k quality, fine art style of greg rokowski, and art nouveau, nature, full shot, symmetrical, Greg Rutkowski, Charlie Bowater, Beeple, Unreal 5, hyperrealistic, dynamic lighting, fantasy art, fine details, surreal engine 5, octane rendering, 8k, , cinematic, 4k, epic Steven Spielberg movie still, sharp focus, emitting diodes, smoke, artillery, sparks, racks, system unit, motherboard, by pascal blanche rutkowski repin artstation hyperrealism painting concept art of detailed character design matte painting, 4 k resolution blade runner, sf, intricate artwork masterpiece, ominous, matte painting movie poster, golden ratio, trending on cgsociety, intricate, epic, trending on artstation, by artgerm, h. r. giger and beksinski, highly detailed, vibrant, production cinematic character render, ultra high quality model
    & {\rotatebox[origin=l]{90}{\textsc{None}}}
    & -
    & \includegraphics[width=0.1\textwidth]{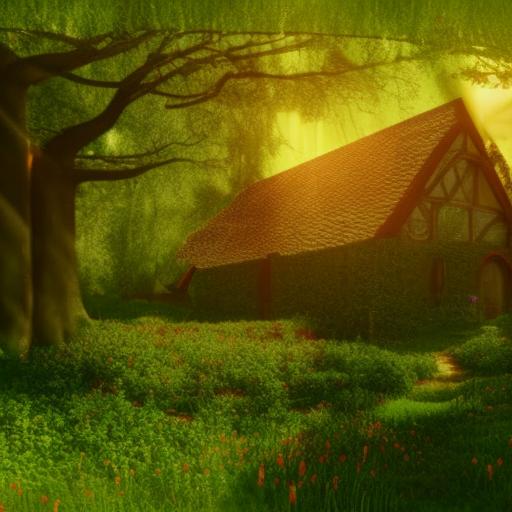} \\
    
    & {\rotatebox[origin=l]{90}{\textsc{\thead{Ground\\Truth}}}}
    & ugly, tiling, poorly drawn hands, poorly drawn feet, poorly drawn face, out of frame, extra limbs, disfigured, deformed, body out of frame, blurry, bad anatomy, blurred, watermark, grainy, signature, cut off, draft
    & \includegraphics[width=0.1\textwidth]{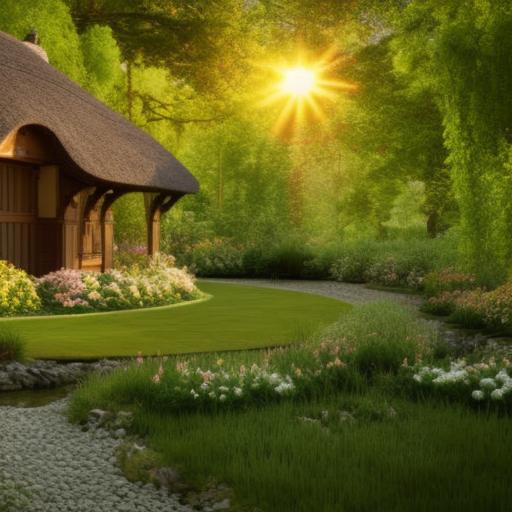} \\
    
    & {\rotatebox[origin=l]{90}{\textsc{Promtist}}}
    & -
    & \includegraphics[width=0.1\textwidth]{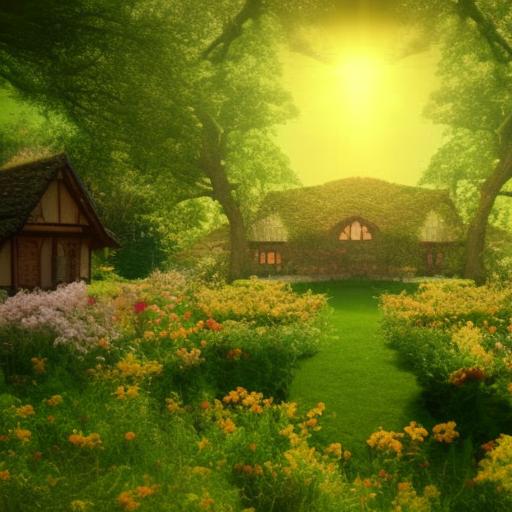} \\

    & {\rotatebox[origin=l]{90}{\textsc{RL}-only}}
    & , artillery, sparks, racks, motherboard, by pascal blanche rutkowski,, cinematic, 4k, epic Steven Spielberg movie still, sharp focus, emitting diodes, smoke, artillery, sparks, racks, system unit, motherboard... %, by pascal blanche rutkowski,.
    & \includegraphics[width=0.1\textwidth]{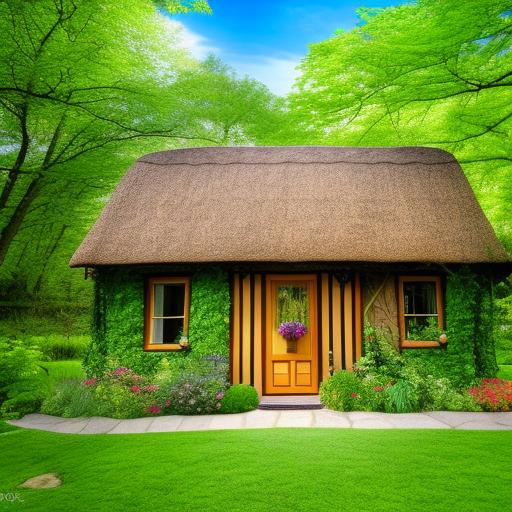} \\
    
    & {\rotatebox[origin=l]{90}{\textsc{SFT}-only}}
    & elongated body, cropped image, red eyes, out of frame, draft, deformed hands, signatures, big hair, twisted fingers, double image, long neck, malformed hands, multiple heads, extra limb, ugly, poorly drawn hands... %, missing limb, disfigured, cut-off, kitsch, ugly, over saturated, grain, low-res, Deformed, blurry, bad anatomy, disfigured, poorly drawn face, mutation, mutated, floating limbs, disconnected limbs, out of focus, long body, disgusting, poorly drawn,
    & \includegraphics[width=0.1\textwidth]{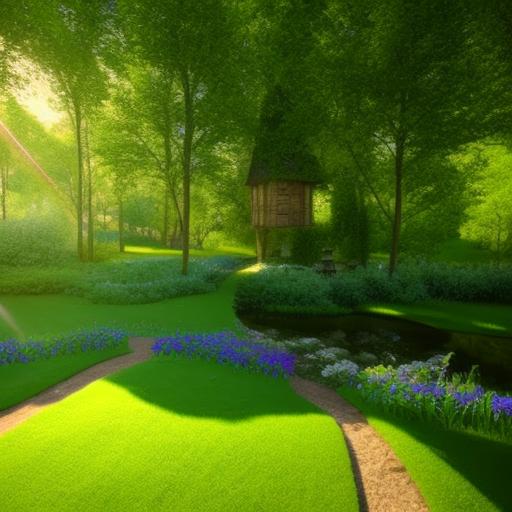} \\
    
    & {\rotatebox[origin=l]{90}{\textsc{SFT+RL}}}
    & elongated body, cropped image, red eyes, out of frame, draft, deformed hands, signatures, big hair, twisted fingers, double image, long neck, malformed hands, multiple heads, extra limb, ugly, poorly drawn hands... %, missing limb, disfigured, cut-off, kitsch, ugly, over saturated, grain, low-res, Deformed, blurry, bad anatomy, disfigured, poorly drawn face, mutation, mutated, floating limbs, disconnected limbs, out of focus, long body, disgusting, poorly drawn,
    & \includegraphics[width=0.1\textwidth]{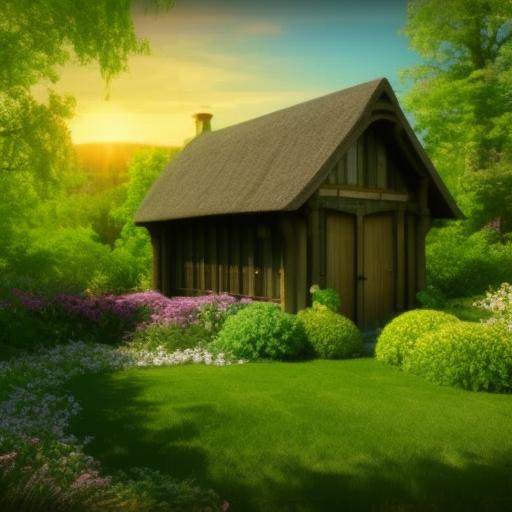} \\
    
    \bottomrule
  \end{tabular}
  \caption{NegOpt compared to all baselines and methods.}
  \label{table:appendix_examples_1}
\end{table*}

\end{document}